\documentclass[english]{article}
\usepackage[T1]{fontenc}
\usepackage[latin9]{inputenc}
\usepackage{geometry}
\usepackage{babel}
\usepackage{float}
\usepackage{amsbsy}
\usepackage{amstext}
\usepackage{graphicx}
\usepackage[unicode=true,pdfusetitle,
 bookmarks=true,bookmarksnumbered=false,bookmarksopen=false,
 breaklinks=false,pdfborder={0 0 1},backref=false,colorlinks=false]
 {hyperref}

\makeatletter
\newcommand{\lyxaddress}[1]{
	\par {\raggedright #1
	\vspace{1.4em}
	\noindent\par}
}

\makeatother

\begin{document}
\title{Faster Biological Gradient Descent Learning}
\author{Ho Ling Li$^{1}$}
\maketitle

\lyxaddress{$^{1}$School of Psychology\\
 University of Nottingham, Nottingham NG7 2RD\\
 U.K.}
\begin{abstract}
Back-propagation is a popular machine learning algorithm that uses
gradient descent in training neural networks for supervised learning,
but can be very slow. A number of algorithms have been developed to
speed up convergence and improve robustness of the learning. However,
they are complicated to implement biologically as they require information
from previous updates. Inspired by synaptic competition in biology,
we have come up with a simple and local gradient descent optimization
algorithm that can reduce training time, with no demand on past details.
Our algorithm, named dynamic learning rate (DLR), works similarly
to the traditional gradient descent used in back-propagation, except
that instead of having a uniform learning rate across all synapses,
the learning rate depends on the current neuronal connection weights.
Our algorithm is found to speed up learning, particularly for small
networks.
\end{abstract}

\section*{Introduction}

In the past decades, back-propagation has become the go-to machine
learning algorithm in training neural networks. It utilizes stochastic
gradient descent (SGD) to minimize a cost function $C$ by adjusting
the connection weights $w_{ij}$ with $\Delta w_{ij}=-\eta\frac{\partial C}{\partial w_{ij}}$
at each time-step. However, learning with the SGD orignally purposed
in back-propagation can be very slow. In addition, networks have higher
risk to overfit when training is slow \cite{Hardt15}. Therefore,
algorithms have been developed to speed up convergence and improve
robustness of learning. These algorithms are mainly separated into
two categories: SGD and adaptive learning. Examples of adaptive learning
algorithms are Adagrad \cite{Duchi11}, RMSprop \cite{Hinton12},
and Adam \cite{Kingma15}. They are usually faster than SGD algorithms,
such as momentum \cite{Plaut86} and Nesterov momentum \cite{Nesterov83}.
On the other hand, networks trained with SGD are better at generalization
than with adaptive learning \cite{Wilson17}. In order to benefit
from both the training speed and generalization capability, several
algorithms have been designed to transit from adaptive learning to
SGD during training \cite{Keskar17,Luo19}. Regardless of which categories
these algorithms belong, they require combining past updates with
the current weight update, which make them complicated to implement
biologically.

Inspired by synaptic competition in biology, we have come up with
a simple and local gradient descent optimization algorithm that encourages
the potentiation of strong synapses and suppresses the growth of weak
synapses. This algorithm, named dynamic learning rate (DLR), works
similarly to the traditional gradient descent used in back-propagation,
except that instead of having a uniform learning rate across all synapses,
the learning rate depends on the current connection weights of individual
synapses and the $L_{2}$ norm of the weights of each neuron. It is
found to speed up learning, particularly for small networks, with
no demand on past information, hence making it biologically plausible.

\section*{Results}

The design of DLR is based on the ideas that synaptic transmissions
are metabolically expensive, thus pushing neurons to lower the number
of strong synapses to save energy \cite{Harris2012,Howarth12,Knoblauch10,Levy1996}.
Unlike the traditional SGD that uses the same learning rate $\eta$
for all synapses, to quicken the rise of strong synapses and speed
up the diversification of the connection strength between neurons,
DLR encourages neurons to form strong connections to a handful of
neurons of their neighbouring layers by assigning higher learning
rate ($\eta_{ij}$) to synapses with bigger weights ($w_{ij}$):

\begin{equation}
\eta_{ij}=-\eta_{0}\,\frac{|w_{ij}|+\alpha}{||\boldsymbol{w_{j}}||+\alpha}\label{eq:DLR_eta}
\end{equation}

\begin{equation}
\Delta w_{ij}=-\eta_{ij}\,\frac{\partial C}{\partial w_{ij}},\label{eq:DLR}
\end{equation}
where $i$ represents the indices of the post-synaptic neurons and
$j$ represents the indices of the pre-synaptic neurons. The parameter
$\alpha$ is set at the range of values such that at the beginning
of training $\alpha>||\boldsymbol{w_{j}}||\gg w_{ij}$ so that all
synapses have similar learning rate. As learning progresses, the learning
rate of all synapses decreases, but large synapses would retain a
relatively large learning rate while the learning rate of small synapses
would become small. Here, $||\boldsymbol{w_{j}}||$ is summing over
all the post-synaptic weights of a pre-synaptic neuron, leading to
each pre-synaptic neuron having strong connections to a limited amount
of post-synaptic neurons only. However, DLR also works by replacing
this term with $||\boldsymbol{w_{i}}||$, hence

\begin{equation}
\eta_{ij}=-\eta_{0}\,\frac{|w_{ij}|+\alpha}{||\boldsymbol{w_{i}}||+\alpha},\label{eq:DLR_eta_v2}
\end{equation}
which promotes every post-synaptic neuron to form strong connections
to a subset of pre-synaptic neurons instead. Whether Eq. \ref{eq:DLR_eta}
or Eq. \ref{eq:DLR_eta_v2} would perform better depends on the network
architecture. Since most networks tend to have decreasing numbers
of neurons for deeper layers, Eq. \ref{eq:DLR_eta} is more applicable
in general. We note that the proposed modulation of learning can easily
be imagined to occur in biology, as it only requires each neuron to
know the connection strength with its own pre- or post-synaptic neurons.

\subsection*{Training speed compared to standard methods}

\begin{figure}[h]
\includegraphics[scale=0.45]{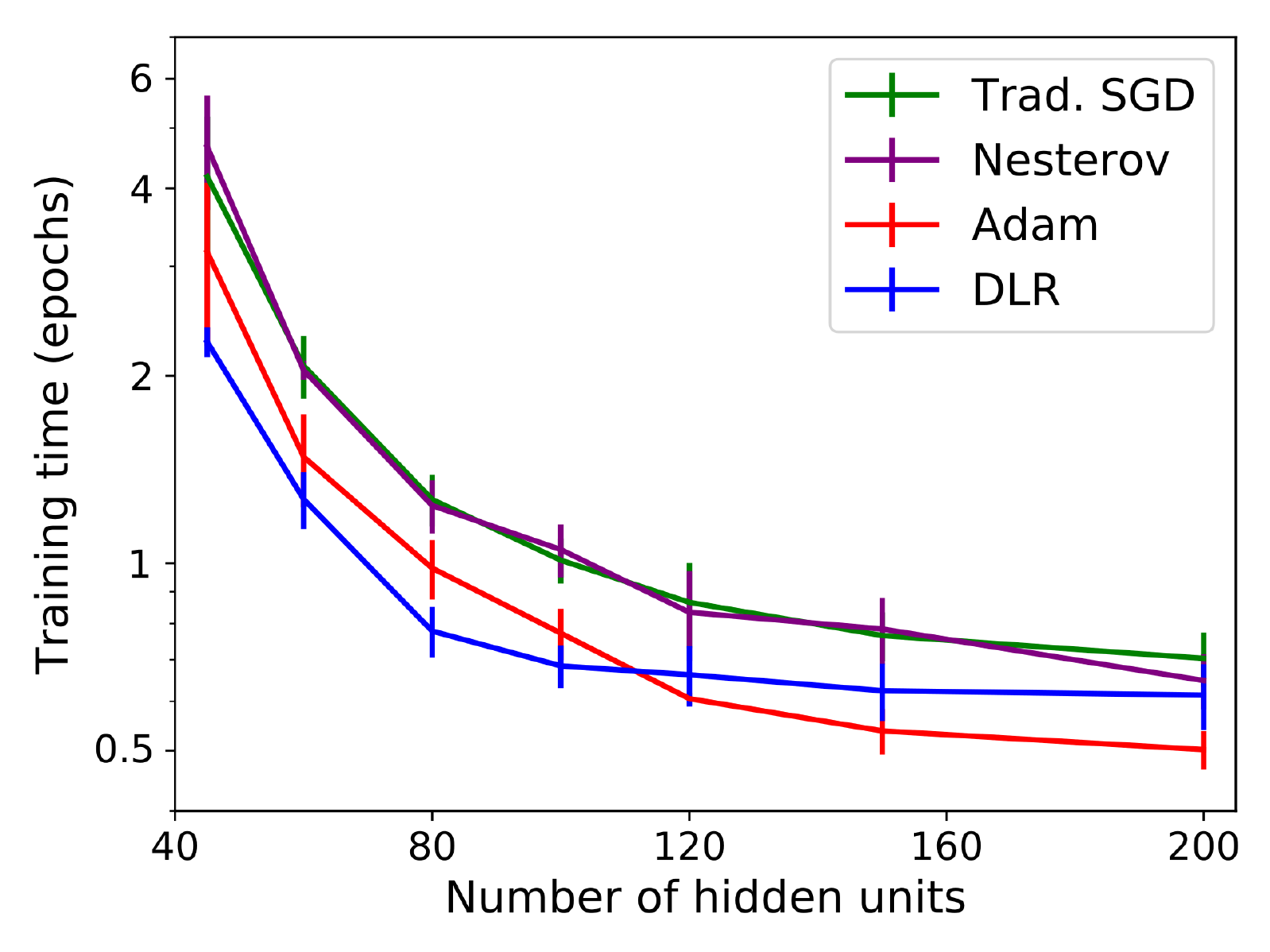}\includegraphics[scale=0.45]{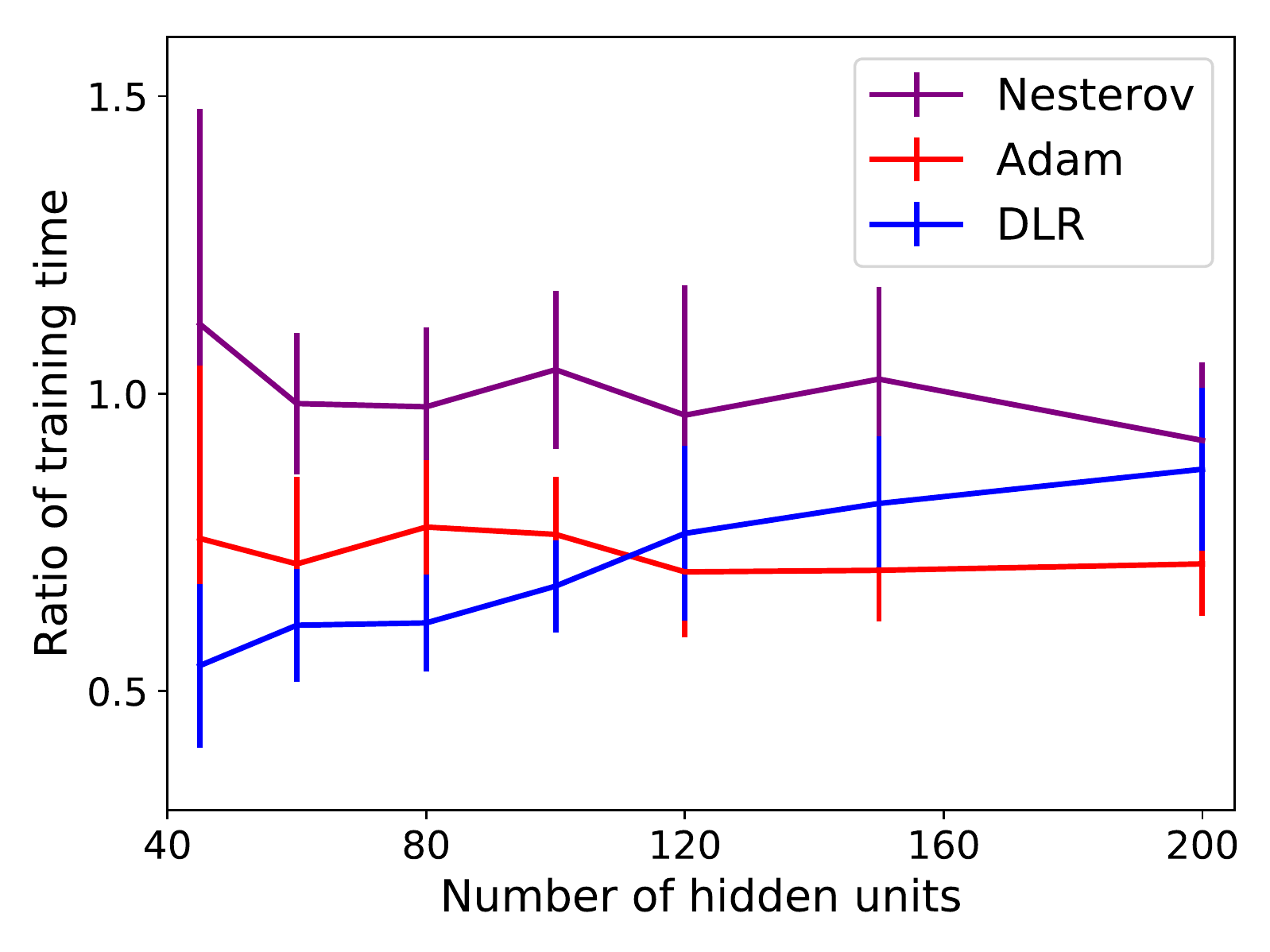}

\caption{\textbf{Comparisons of training speed between traditional SGD, Nesterov
momentum, Adam and DLR}. The networks with back-propagation are trained
with one of the four algorithms. The network size is varied by adjusting
the number of neurons in the hidden layer. The left figure shows the
training time of the four algorithms. To illustrate the change in
training time by switching from traditional SGD to the other three
algorithms, the ratio of their training time to that of traditional
SGD is displayed on the right. Networks equipped with DLR reach the
designated accuracy the fastest when the networks have fewer than
$100$ hidden units. The uncertainties are standard deviations across
multiple runs. \label{fig:training_speed}}
\end{figure}

To test the performance of DLR, we implement a multi-layer network
trained with back-propagation with one hidden layer to classify hand-written
digits from the MNIST dataset to a benchmark accuracy of $96\%$.
We compare the performance of DLR with the traditional SGD in back-propagation,
Nesterov momentum \cite{Nesterov83}, and the commonly used adaptive
learning algorithm Adam \cite{Kingma15}. All the algorithms involve
one or two parameters except Adam, which involves four parameters:
$\alpha$, $\beta_{1}$, $\beta_{2}$ and $\epsilon$. From a brief
check, training becomes faster after changing $\alpha$ and $\epsilon$
from the default values suggested by Adam's authors Kingma and Ba
\cite{Kingma15}. Therefore, Adam is optimized with respect to $\alpha$
and $\epsilon$. The other three algorithms have all of their parameters
optimized. Networks with fewer than $100$ hidden units train the
fastest with DLR though Adam is not significantly slower than DLR,
Figure \ref{fig:training_speed}. As the network size increases, Adam
performs the best.

\subsection*{Robust for small networks}

\begin{figure}[H]
\includegraphics[scale=0.6]{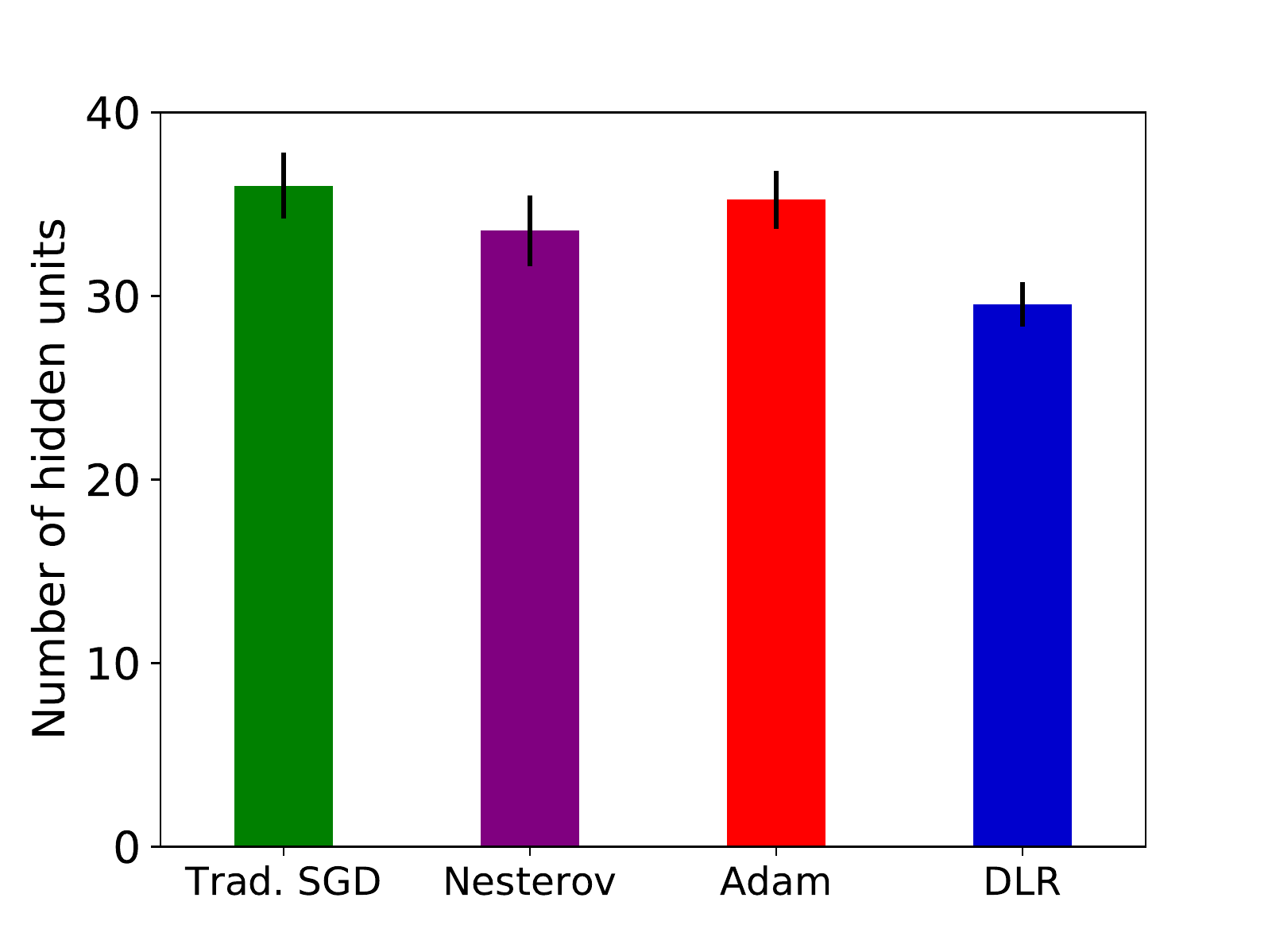}

\caption{\textbf{Mimimal requirement on network size}. By gradually reducing
the number of neurons in the hidden layer, network sizes are decresaed
until the networks fail to reach an accuracy level of $96\%$. Networks
trained with DLR needs on average fewer neurons than the other three
algorithms. The uncertainties are standard deviations across multiple
runs. \label{fig:small_networks}}
\end{figure}

Next, we are interested in knowing if DLR is more robust for small
networks compared to other algorithms, i.e. whether DRL allows small
networks to reach the designated accuracy that would otherwise not
be able to reach if they are trained with other learning algorithms.
To test it, we reduce the number of neurons at the hidden layer until
the networks are no longer able to converge. Figure \ref{fig:small_networks}
shows the minimal network size that the network can still satisfy
the accuracy requirement. Since the change in the network architecture
may require different parameter values for the algorithms, we scan
the performance of each algorithm across a parameter space. Compared
to traditional SGD, Adam, and Nesterov momentum, which on average
demands $36.3\pm1.2$, $35.3\pm1.6$ and $33.7\pm2.0$ hidden neurons
respectively, networks trained with DLR only require $29.7\pm1.6$
neurons to reach $96\%$ accuracy. Our speculated explanation is because
DRL allows large weights to have high learning rates, in the case
of networks being stuck at a local minimum, undesirable big weights
can depress more quickly hence releasing the networks from that local
minimum.

\subsection*{Synapse-specific learning rate benefits learning}

\begin{figure}[H]
\includegraphics[scale=0.6]{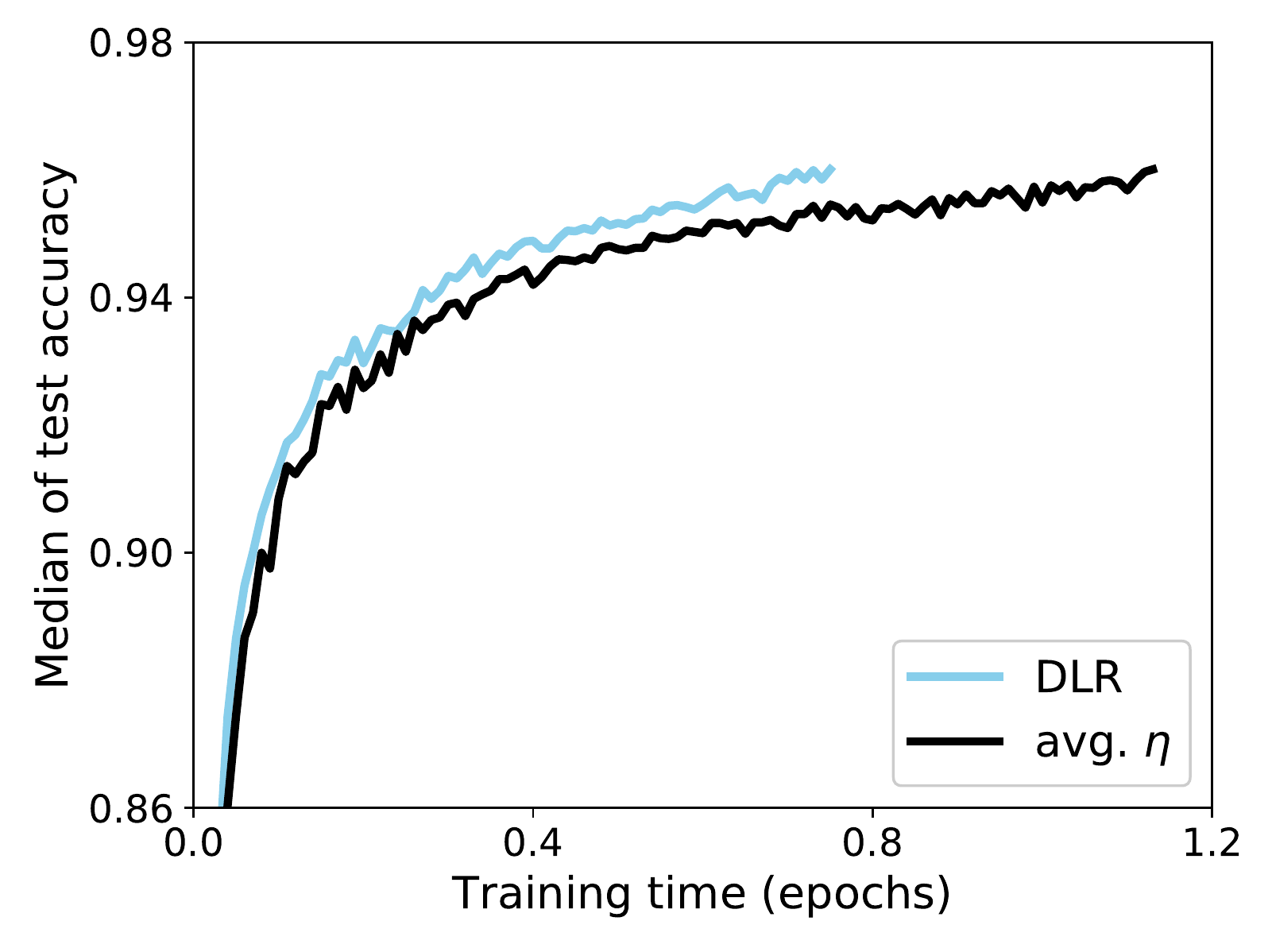}

\caption{\textbf{Median of test accuracy over training time}. The median of
test accuracy over multiple runs is shown until it reaches the designated
accuracy. Networks equipped with DLR finish training within $0.8$
epoch in majority of the runs. Networks trained with the average learning
rate of DLR take at least $1.1$ epochs in most of the runs. \label{fig:annealing}}
\end{figure}

During training, when implemented with DLR, the learning rate of most
synapses would drop. To check that the improvement of training speed
is not primarily due to a gradual global decrease in learning rate
but instead due to the learning rate being synapse-specific, we have
compared the learning time of networks trained with DLR with networks
trained with the average learning rate of DLR. This is achieved by
measuring the average learning rate between the input and hidden layers,
and between the hidden and output layers of the networks trained with
DLR, then fitting the average learning rate with $a\,\textrm{exp}(bt^{1/3}+ct)+d$,
where $t$ represents the training time. The fits are implemented
into new networks, which have to learn to classify the MNIST dataset
with that predetermined learning rate. Networks learnt with DLR reaches
$96\%$ accuracy with $(0.77\pm0.10)$ epoch in contrast to $(1.14\pm0.22)$
epochs when networks learnt with the average learning rate, showing
the criticality of the learning rate being synapse-specific. Figure
\ref{fig:annealing} shows the median of test accuracy over multiple
runs, demonstrating how networks with DLR progress faster during training.

\section*{Discussion}

DLR has shown the possibility of a local and biological gradient descent
optimization algorithm that can speed up neural network training with
back-propagation. It only requires online information, which may have
the benefits of lower memory usage at large networks compared to algorithms
that require storing information from past updates.

It is shown to have comparable training speed with the popular adaptive
learning algorithm Adam for networks with small and medium sizes,
i.e. when the parameters in the networks are not redundant. In addition,
DLR is found to be more robust than traditional SGD, Nesterov momentum
and Adam as it allows small networks to acquire an accuracy level
that otherwise would not be able to achieve. It shows that DLR can
find the solution more efficiently even when the number of weight
parameters is restrained. Here, we have conducted the tests on MNIST
dataset with networks that are small compared to deep networks that
are used to categorize much more complicated images. Therefore, it
is uncertain how DLR will perform in deep networks. We wonder if it
is also applicable in those networks, which even though have significantly
larger network sizes, may still suffer the issue of insufficient weight
parameters. On the other hand, DLR may allow the use of smaller deep
networks by guaranteeing similar performance as the larger ones, and
provide the benefits of less computation time and memory.

In recent years, many algorithms that can effectively speed up learning
are adaptive learning algorithms, which are found to have not as good
generalization capability as SGD \cite{Wilson17}. It would be interesting
to test if DLR, as a SGD, performs well in generalization while still
has comparable training speed as adaptive learning algorithms.

\section*{Methods}

We use networks with one hidden layer, logistic units without bias,
and one-hot encoding at the output. Weights are updated according
to the mean squared error back-propagation rule without regularization.

\section*{Acknowledgments}

This project is supported by the Leverhulme Trust with grant number
RPG-2017-404. We would also like to thank University of Nottingham
High Performance Computing for providing computing powers for this
research.

\bibliographystyle{pnas-new}

\end{document}